\documentclass[journal]{IEEEtran}

\usepackage{graphicx}
\usepackage{subfigure}
\usepackage{booktabs}

\ifCLASSINFOpdf
\else
\fi

\begin{document}

\title{Short-term Load Forecasting with Dense Average Network}

\author{Zhifang Liao,~\IEEEmembership{Member,~IEEE,}
        Haihui Pan,
        Qi Zeng, 
        Xiaoping Fan,~\IEEEmembership{Member,~IEEE,}
        Yan Zhang,
        and \\ Song Yu,~\IEEEmembership{Member,~IEEE,}


\thanks{Zhifang Liao, Haihui Pan, Qi Zeng, Xiaoping Fan and Song Yu are with the School of Computer Science and Engineering, Central South University, Changsha, 410083, P. R. of China.}
\thanks{Yan Zhang is with School of Computing, Engineering and Built Environment, Department of Computing, Caledonian University, UK}
\thanks{(Corresponding author email: xpfan@mail.csu.edu.cn).}}

\markboth{}
{Shell \MakeLowercase{\textit{et al.}}: Bare Demo of IEEEtran.cls for IEEE Journals}
\maketitle

\begin{abstract}

As an important part of the power system, power load forecasting directly affects the national economy. The data shows that improving the load forecasting accuracy by 0.01\% can save millions of dollars for the power industry. Therefore, improving the accuracy of power load forecasting has always been the pursuing goals for a power system. Based on this goal, this paper proposes a novel connection, the dense average connection, in which the outputs of all preceding layers are averaged as the input of the next layer in a feed-forward fashion. Based on dense average connection , we construct the dense average network for power load forecasting. The predictions of the proposed model for two public datasets are better than those of existing methods. On this basis, we use the ensemble method to further improve the accuracy of the model. To verify the reliability of the model predictions, the robustness is analyzed and verified by adding input disturbances. The experimental results show that the proposed model is effective and robust for power load forecasting.
\end{abstract}


\IEEEpeerreviewmaketitle

\section{Introduction}
\IEEEPARstart{L}{oad} forecasting plays a key role in the management and dispatching of power systems. Load forecasting involves forecasting the load demand of a future time span. Load forecasting within an interval of an hour to a week is often referred to as short-term load forecasting (STLF). The accuracy of STLF significantly affects the economic operation and reliability of a power system, and inadequate STLF may lead to insufficient reserve capacity and allocation with expensive peaking units or cause unnecessarily large reserve capacity. Both these outcomes are related to increased operating costs. Therefore, accurate load forecasting plays an important role in energy market analysis and economic dispatch in the power industry. In future smart grids, reliable STLF is of great significance for operators to manage grids with higher efficiency and lower cost.

Load demand is a non-stationary process that is affected by many factors, including weather conditions, seasonal effects, socioeconomic factors, and random effects \cite{1-hahn2009electric}, which makes load demand difficult to predict. At present, many methods have been proposed for STFL. Most of these methods are based on statistical methods or artificial intelligence algorithms. In the early days, the autoregressive moving average model (ARMA) \cite{2-huang2003short}, fuzzy logic \cite{3-rejc2011short}, expert systems \cite{4-kandil2002long} and other algorithms were widely used in load forecasting. In recent years, artificial intelligence methods such as neural networks and support vector machines \cite{5-de2011short,30-ceperic2013strategy} have been proposed. Some different types and variants of neural networks have also been proposed and applied to STLF, such as wavelet neural networks \cite{9-guan2012very,8-chen2009short} , extreme learning machines (ELMs) \cite{10-li2015short} and wavelet-based hybrid neural
networks \cite{34-li2015novel}. At present deep neural networks (DNNs) have facilitated great achievements in many fields \cite{11-bahdanau2014neural,12-tompson2015efficient}. For the success of deep neural networks, model structure design and model depth play an important role \cite{19-szegedy2015going,20-hu2018squeeze,15-he2016deep}. The application of DNNs to short-term load forecasting is a relatively new topic. In \cite{16-ryu2017deep,17-chen2018short}, deep artificial neural networks and deep residual networks are applied to load forecasting. In \cite{35-kong2017short}, long short-term memory (LSTM) recurrent neural network-based is applied to the task of short-term load forecasting for individual residential households. In \cite{36-pramono2019deep}, a wavenet based model that employs dilated causal residual convolutional neural network (CNN) and LSTM layer is applied to load forecasting.

In this work, we propose a novel model structure for load forecasting. First, we propose the dense average connection, in which the outputs of all preceding layers are averaged as the input of the next layer in a feed-forward fashion. Based on the dense average connection, we build the dense average network. Second, we further improve the prediction accuracy by the ensemble method. Finally, we perturb the input of the model to varying degrees to verify the robustness of the model. The main contributions of this work are as follows:

\begin{itemize}
    \item We propose the dense average connection, and based on this connection, we build the dense average network. On two public datasets, we evaluate the validity of the dense average network. The dense average network does not require external feature extraction and only uses the data of load, temperature and date information as input.
    
    \item We use the ensemble method to further improve the prediction effect. The experimental results show that compared with a single model, the ensemble method can not only improve the prediction accuracy but also reduce the standard deviation and peak value of the final prediction bias.
    
    \item To ensure the reliability of model prediction, we conduct a comprehensive analysis on the robustness of the model. We disturb the original load data and temperature data to different degrees. The experimental results show that the proposed model is very robust to data noise.
\end{itemize}
The remainder of this paper is organized as follows. In section II, we introduced the model structure, ensemble method, and implementation details. In section III, We compare the proposed model with the current methods on two public datasets to verify the validity of the proposed model. We will also apply the proposed model to a real dataset.Section IV summarizes the paper and puts forward the future work. We will release our experimental code and trained models later.

\section{Methodology}
This paper proposes the Dense Average Network (DaNet) for short-term load forecasting. We first construct features as input to the model from three aspects: historical load data, historical temperature data, and date information of historical load data. Secondly, we introduce the origin of the Dense Average connection. Based on the Dense Average connection, we build the Dense Average Network. After that, we use ensemble method to improve the accuracy of the load forecasting. Finally, in order to ensure the reliability of the model, we performed a robust analysis on the model.

\subsection{Model Input Variables }
The actual load demand is often affected by many factors, such as the economy, weather, seasons, and holidays \cite{1-hahn2009electric}. Therefore, the features that are selected as the input of the model greatly influence the final prediction result. Meanwhile, we need to note that the raw data of the model input variables we construct should be easily accessible so that our method can be applied in most real-world scenarios. In this work, we mainly construct features from three aspects: historical load data, historical temperature data and date information. Specifically, the variables related to the input are listed in Table I.

For the historical load data, we extract historical data from the past two days. Since the load data interval size of the public datasets is 1 hour, the data size of the load data is 48. The recent fluctuations of data can often indicate the recent trends of data. For example, if the load data trend upward, it is likely that the load data also trend upward in the near future. Therefore, to obtain the recent fluctuations of data, we extract the slope of the historical load data. The slope value {\em $S_{h}$} at time {\em h} is defined as
\begin{equation}
  S_{h}=\left (L^{h}-L^{h-1}  \right )/\left (h-\left ( h-1 \right )  \right )=L^{h}-L^{h-1}
\end{equation}
where {\em $L^{h}$} is the load value at time {\em h} and {\em $L^{h-1}$} is the load value at time {\em h-1}. When {\em $S_{h}$} \textgreater0, the load value trends upward; when {\em $S_{h}$} \textless0, the load value trends downward. For the historical temperature data, we obtain the temperature data corresponding to the historical load data. Therefore, the data size of the temperature data is also 48. For the date information, we mainly extract two features: month and weekday. We do not extract the two features of season and holiday because the information of season is hidden in the information of month, and the information of holiday is hidden in the information of week. In the experiment, we find that an increase in season and holiday does not increase the accuracy of the model prediction. In data processing, we perform one-hot processing for month and weekday.

\subsection{Dense Average Network Structure}
In \cite{18-huang2017densely}, a novel connection is proposed for image recognition. Layer $\ell$ concatenates the output of all preceding layers (feature maps of the same size), which aims to achieve feature reuse and improve the information flow between layers. Let {\em $H_{\ell}(.)$} be the nonlinear transformation of layer $\ell$, then the output of layer $\ell$ is
\begin{equation}
  x_{\ell}=H_{\ell}([x_{0},x_{1},...,x_{\ell-1}])
\end{equation}
where [$x_{0}$,$x_{1}$,...,$x_{\ell-1}$] indicates that the input and feature maps from layer 1 to layer $x_{\ell-1}$ are concatenated according to depth. Based on the convolution operation, DenseNet \cite{18-huang2017densely} can obtain superior results when processing two-dimensional data with spatial correlations such as images. However, because the load data is one-dimensional, it is impossible to use DenseNet directly for load forecasting. Of course, we can also directly concatenate the network output of all processing layers to form a large vector, but there are two major problems with this method. First, the parameters of the network model increase dramatically due to the concatenating of all processing layers. Second, as the network depth increases, this concatenating method causes the model to fail to train.We verify this in experiments. Therefore, a feasible idea is to use a combination method to keep the parameters of the {\em$x_{\ell}$} layer the same as those of the processing {\em $x_{\ell-1}$} layer.

\begin{table}[]
\centering
\caption{Model input variables and input related variables}
\begin{tabular}{@{}lrl@{}}
\toprule
Input Variable & \multicolumn{1}{l}{Size} & Description of Input Variable \\ \midrule
L & 48 & Load data for the last two days \\
T & 48 & Temperature of load data \\
S & 48 & Slope of load data \\
$L_{i}$ & 1 & The {\em i}-th element of L \\
$S_{i}$ & 1 & The {\em i}-th element of S \\
L\_S & 96 & {[}{[}$S_{1}$,$L_{1}${]},..,{[}$S_{48}$ ,$L_{48}${]}{]} \\
W & 7 & One-hot code for weekday \\
M & 12 & One-hot code for Month \\ \bottomrule
\end{tabular}
\label{tab:table1}
\end{table}

The first method we use is to add all the outputs of the processing layer. Unfortunately, this method can cause problems with gradient explosion. We analyze the essential problem of this method, which provides the idea for the method we ultimately adopt. Let $x_{0}$ be the input of the model, and the output of the model with $\ell$ layers is
\begin{equation}
\centering
  x_{\ell}=H_{\ell}(x_{\ell-1})+\sum _{i=0}^{\ell-1}x_{i}
\end{equation}
The total loss of the neural network for the back propagation of {\em$x_{0}$} is calculated as
\begin{equation}
  \frac{\partial L}{\partial x_{0}} = \frac{\partial L}{\partial x_{\ell}}\frac{\partial x_{\ell}}{\partial x_{0}}\\ =\frac{\partial L}{\partial x_{\ell}}(\frac{\partial H_{\ell-1}(x_{\ell-1})}{\partial x_{0}}+\sum _{i=1}^{\ell-1}\frac{\partial x_{i}}{\partial x_{0}}+1)
\end{equation}
where {\em L} is the loss function of the neural network. Since the outputs of all processing layers are combined by using the addition operation, $\partial x_{i}/\partial x_{0}$ $>$1. Therefore, as the number of layers increases,  $\sum _{i=1}^{\ell-1}\frac{\partial x_{i}}{\partial x_{0}}$ increases linearly. Therefore, when building a deep model, gradient explosion occurs.

We finally use the average operation to combine the outputs of all processing layers.The output of layer ${\ell}$ using this method is
\begin{equation}
x_{\ell}=\frac{1}{\ell+1}(H_{\ell}(x_{\ell-1})+\sum _{i=0}^{\ell-1}x_{i})
\end{equation}
There are three benefits to using the average operation to combine the outputs of all processing layers. First, since the outputs of all processing ${\ell-1}$ layers are averaged, the problem of gradient explosion is largely dealt with, which allows the model to be trained deeply. Second, using the average operation to combine the outputs of all processing layers does not introduce new parameters. Finally, assuming that the distribution of the processing ${\ell-1}$ layers is similar, the output of the processing ${\ell-1}$ layer is combined using the average operation as the input of the second layer, and the output of the layer ${\ell}$ maintains an approximate distribution in the processing layer. We call this connection the dense average connection. To facilitate the establishment of a deeper network, we construct a dense average block. As shown in Figure 1, each dense average block has four fully connected layers. For building a deep network, we simply stack multiple dense average blocks.

Based on the dense average block, we build the dense average network. The structure of the dense average network is shown in Figure 2. Our model has a total of 3 inputs, which are load data, temperature data and date information data. Among them, the dimensions of the temperature data and date data are one-dimensional, so we only use the fully connected layer for feature extraction. The combination of load data and slope data are two-dimensional data, so we can use two-dimensional convolution to extract features. To fully extract features of different scales, we use the design idea of Inception \cite{19-szegedy2015going} for reference. We use four convolution kernels of different sizes to extract features from the load data. The sizes of the four convolution kernels are $1\times 2,2\times 2,3\times 2,4\times 2$, and the step size is set to 1 in all convolution operations. To pay more attention to features with rich information and suppress features with less information, a squeeze-and-excitation (SE) block was proposed in \cite{20-hu2018squeeze} for feature recalibration. This method can play a better role in convolution operations. The structure of the SE block is shown in Figure 3. The overall information of the feature map is obtained through the average pooling operation. After that, two more hidden layers are used to generate weights. Finally, the generated weights are multiplied by the feature maps. In our model, we also add the SE block operation for each convolution layer.
\begin{figure}[]
\centering
\includegraphics[ width=0.13\textwidth]{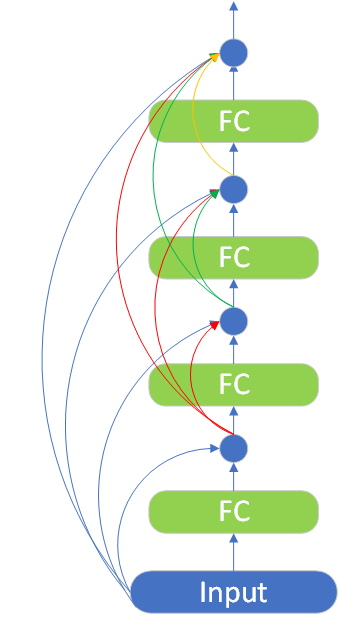}
\caption{Structural diagram of the dense average block. Each dense average block consists of four fully connected layers. The blue circle in the figure represents the average operation.}
\end{figure}

The dense average network uses a total of 5 dense average blocks, and the depth of the model is 22 layers (excluding the input layer and output layer). The number of neurons or kernels in all hidden layers is set to 128. Except for the activation function of the last layer in the SE block, which is sigmoid, the remaining activation functions are set to ReLU \cite{21-dahl2013improving}. The forms of the ReLU and sigmoid are shown in (6) and (7).
{\setlength\arraycolsep{2pt}
\begin{eqnarray}
ReLU(x)&=&max\left \{0,x  \right \}
\\
Sigmoid(x)&=&\frac{1}{1+e^{-x}}
\end{eqnarray}
%

\begin{figure}[]
\centering
\includegraphics[ width=0.35\textwidth]{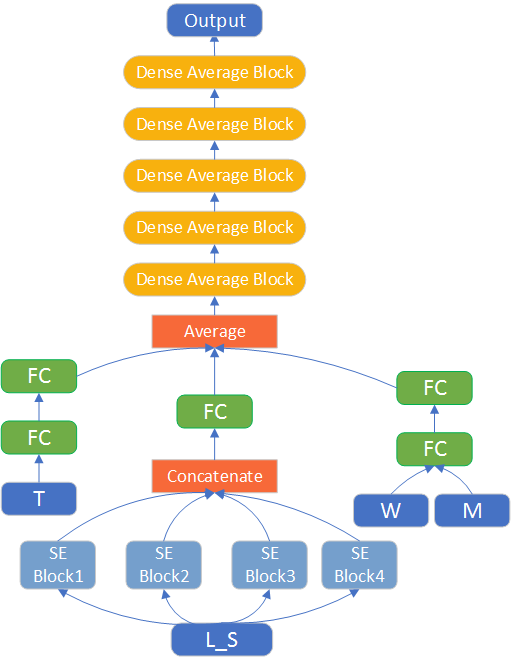}
\caption{Model structure of the dense average network. Model input variables are all listed in Table I.}
\end{figure}
%

\begin{figure}[]
\centering
\includegraphics[ width=0.16\textwidth]{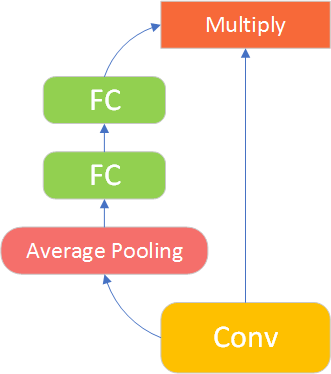}
\caption{ Structure of the squeeze-and-excitation (SE) block.}
\end{figure}


\subsection{Ensemble Method Based on Dense Average Network}
In the field of machine learning, a common approach to improve the prediction results of models is to use the ensemble method based on multiple models. In this work, we first train multiple different dense average networks, then we average the predicted result of each model to obtain the final results. This ensemble method is called Bagging \cite{17-chen2018short}. In fact, the error of the model can be divided into two parts: bias and variance. Bagging reduces the error of the model by lowering the variance error of the model. We assume that {\em f(x)} is the designed model,  {\em f(x;D)} is the prediction of the model on dataset {\em D}, and {\em y} is the label of the sample; then, the expectation for the mean square error of the model prediction is
{\setlength\arraycolsep{2pt}
\begin{eqnarray}
E[(f(x;D)-y)^{2}] &=& E[(f(x;D)-E[f(x;D)])^{2}]+
 \nonumber\\
&+&(E[f(x;D)]-y)^{2}
\end{eqnarray}

Suppose that we have m models, where the error of each model for each sample is {\em $e_{i}$}, the error obeys the multidimensional normal distribution of the zero mean, {\em $E[e^{2}_{i}]=v$} is the variance, and {\em $E[e_{i}e_{j}]=c$} is the covariance. Then, the average error expectation of the ensemble model prediction is
{\setlength\arraycolsep{2pt}
\begin{eqnarray}
E[(\frac{1}{k}\sum_{i=1}^{m}e_{i})^{2}] &=&\frac{1}{k^{2}}E[\sum_{i=1}^{m}(e_{i}^{2}+\sum_{i\neq j}e_{i}e_{j})] 
\nonumber\\
&=&\frac{1}{k}v+\frac{k-1}{k}c
\end{eqnarray}

When the errors of the multiple models trained on the samples are consistent, that is, {\em c=v}, the mean square error of the ensemble model is still {\em v}. When the model's error for the sample is completely irrelevant, that is, {\em c = 0}, the mean square error of the ensemble model is reduced to {\em v/k}, and the model's error decreases linearly with the scale of the ensemble model. Therefore, if bagging can be expected to produce good results, the premise is that the error of a single model should be as small as possible and the difference between models may be significant.

In all ensemble models, we train different models by randomly selecting 90\% of the training set. In the experiment, we find that only 80\% of the training set has a great impact on the performance of a single model. The number of ensemble models is 5. In section 3.3, we specifically discuss the impact of the number of ensemble models on the final results.

\subsection{Implementation Details}
In all experiments, the loss function of the model is set to the mean absolute error, the training batch size is 256, and the optimizer is Adam \cite{23-kingma2014adam}. We set the learning rate schedule for the optimizer. Although the learning rate of each iteration of Adam is self-adaptive, we find that the convergence of the model can be more stable by setting the learning rate schedule in the experiment. Adam's initial learning rate is 0.001, and the learning rate is divided by 10 for every 600 epochs. The total number of training epochs of the model is 1200. The parameters of all the models are initialized by using the truncated normal distribution with a mean value of 0 and a standard deviation of 1. These models are implemented in the Python 3.6 environment using Keras 2.1.0 and TensorFlow 1.9.0 as backends \cite{24-gulli2017deep,25-abadi2016tensorflow}. All experiments are run on a computer with GTX 1080 graphics card. It takes about 40 minutes to train the dense average network with data of two years for 1200 epochs. To evaluate prediction performance, three error indicators are used: mean absolute percentage error (MAPE), mean absolute error (MAE) and root mean squared  error (RMSE).
{\setlength\arraycolsep{2pt}
\begin{eqnarray}
MAPE&=&\frac{1}{N}\sum_{i=1}^{N}\left | \frac{y_{i}-\hat{y}_{i}}{y_{i}} \right |\times 100\%
\\
MAE&=&\frac{1}{N}\sum_{i=1}^{N}\left | y_{i}-\hat{y}_{i} \right |
\\
RMSE&=&\sqrt{\frac{1}{N}\sum_{i=1}^{N}(y_{i}-\hat{y}_{i})^2}
\end{eqnarray}
where {\em N} is the number of samples, $y_{i}$ is the actual load value, and $\hat{y_{i}}$ is the predicted load value.

\section{Experiments}

\subsection{Datasets}
We use two public datasets (the ISO New England (ISO-NE) dataset and North-American Utility (NAU) dataset) to verify the validity of the proposed model. The ISO-NE dataset and NAU dataset both contain load and temperature data with a one-hour resolution. The time range of the ISO-NE dataset is from March 2003 to December 2014 and  the time range of the NAU dataset is from January 1985 to October 1992.

\subsection{Effectiveness of Dense Average Network}
In this case, we mainly analyze the difference between the performance of the model based on the dense average block (dense average network, DaNet) and that of the model based on the fully connected layer (artificial neural network, ANN). The structure of DaNet is shown in Figure 2. We replace all the dense average blocks in Figure 2 with fully connected layers in the ANN structure for comparison. Since the dense average connection does not introduce new training parameters, as long as DaNet and ANN have the same number of layers and the number of neurons in each layer is the same, the parameters for the two models are consistent. For the sake of fairness, both models use the same training method and parameter initialization method.

\begin{figure}
\centering
\includegraphics[ width=0.38\textwidth]{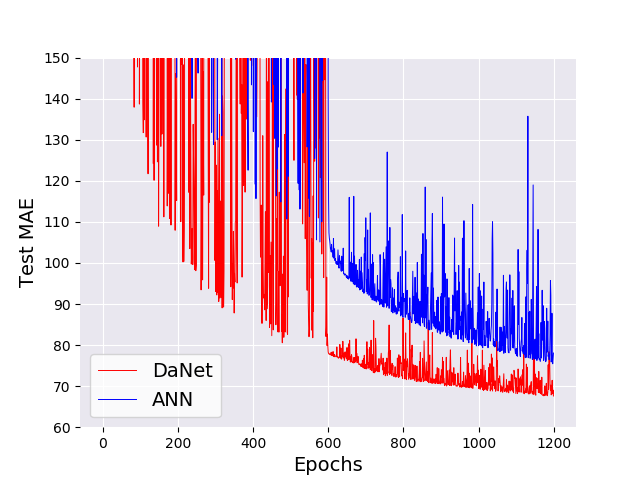}
\caption{Test loss values of dense average network (DaNet) and artificial neural network (ANN) on the ISO-NE dataset. We train each model separately 5 times. The results of the solid line are obtained by averaging the test loss values of the 5 models.}
\end{figure}

\begin{table*}[]
\centering
\caption{MAPE(\%), MAE, MAX, and STD for ensemble methods with different numbers of models. MAX: Maximum prediction bias on the test set; SD: Standard deviation of the prediction bias for the test set.}
\begin{tabular}{@{}lrrrrrrrrrrrr@{}}
\toprule
\multicolumn{13}{c}{Ensemble Size} \\ \midrule
 & \multicolumn{1}{c}{1} & \multicolumn{1}{c}{2} & \multicolumn{1}{c}{3} & \multicolumn{1}{c}{4} & \multicolumn{1}{c}{5} & \multicolumn{1}{c}{6} & \multicolumn{1}{c}{7} & \multicolumn{1}{c}{8} & \multicolumn{1}{c}{9} & \multicolumn{1}{c}{10} & \multicolumn{1}{c}{11} & \multicolumn{1}{c}{12} \\
\hline
MAPE & 0.3842 & 0.3698 & 0.3655 & 0.3658 & \textbf{0.3617} & 0.3655 & 0.3662 & 0.3673 & 0.3691 & 0.3718 & 0.3747 & 0.3765 \\
MAE & 58.93 & 57.46 & 56.69 & 56.62 & \textbf{56.03} & 56.53 & 56.65 & 56.81 & 57.08 & 57.49 & 57.92 & 58.21 \\
MAX & 387.05 & 382.14 & 390.93 & 380.40 & 375.77 & 377.93 & 374.79 & 375.15 & \textbf{372.03} & 372.06 & 374.19 & 375.80 \\
SD & 56.52 & 54.30 & 54.38 & \textbf{53.91} & 54.15 & 54.63 & 54.95 & 55.03 & 55.32 & 55.56 & 55.97 & 56.15 \\ \bottomrule
\end{tabular}
\label{table2}
\end{table*}
We use the ISO-NE dataset for performance comparison of this case. We use the year 2004 in the dataset as the training set and the year 2005 in the dataset as the test set. We divide the last month of the training set into the validation set. We train each model 5 times and average the test loss values to obtain the final result. From Figure 4, we can see that after the number of training epochs reaches 600, the fluctuation range of the loss value of the model on the test set is greatly reduced for both DaNet and ANN. This result shows that when using Adam as an optimizer, the models can have better convergence performance by setting the learning rate schedule. Additionally, after reducing the learning rate, the fluctuation range of the solid red line is significantly lower than that of the solid blue line, which means that DaNet has better convergence properties than ANN. We compare the final test loss of DaNet and ANN more specifically. The MAE is 67 for DaNet and 75 for ANN. Compared to ANN, DaNet reduces the MAE of the test set by 10.7\%. The final results of the experiment show that compared with ANN, DaNet has better prediction results and convergence performance.

\subsection{Ensemble Scheme}
In this case, we focus on the impact of the ensemble size on the prediction effect. Specifically, we focus on two questions: for the bagging ensemble method, what number of models can best predict the results? Additionally, do the standard deviation of the predicted results and the predicted extreme values of the ensemble model differ from those of the individual models?

We focus on the differences between the standard deviation of the predicted results and the extreme deviation of the predicted results because the energy management efficiency of the smart grid may be strongly affected by the peak error, and a predictor with low variance may be favored over a predictor with lower average error but higher peak error. This is because underestimating energy demand can have a negative impact on the demand response, making it more difficult to control overload conditions; overestimating energy, however, can lead to unexpected overproduction. In both cases, the greater the estimation error, the higher the administrative costs involved.

We use the ISO-NE dataset to analyze the performance of the ensemble model. The training set ranges from 2007 to July 2008, and the test set ranges from August 1, 2008 to August 31, 2008. We explore the effect of ensemble size from 1 to 12, the maximum prediction bias and the standard deviation of the prediction bias. The experimental results are shown in Table II. When the ensemble size is 5, the predicted performance is the best. At the same time, the prediction results of all the ensemble models are better than those of the single models. We find that the prediction bias and the maximum prediction bias of the model with lower MAPE or MAE are not the lowest. We also find that the maximum deviation of the prediction when more models are used for integration is significantly lower than that of a single model, which is of great significance for deploying the ensemble model into the actual environment. In the following experiments, all the ensemble sizes of the ensemble model are set to five.

\subsection{Performance of the Proposed Model on the ISO-NE Dataset}
In this use case, we compare the proposed model with existing methods on the ISO-NE dataset. Because some methods choose different test set time ranges, we performed 2 comparisons. We first compare with the existing 3 methods \cite{26-shamsollahi2001neural,9-guan2012very,10-li2015short}. In \cite{26-shamsollahi2001neural}, a prediction method based on ANN is proposed. In \cite{9-guan2012very}, a wavelet neural network method for data prefiltering is proposed. In \cite{10-li2015short}, a short-term load prediction method based on wavelet transform, a limit learning machine and an improved artificial bee colony algorithm is proposed.


The training set is from January 1, 2007 to June 2008, with the last month being used as the validation set.The test set ranges from July 1, 2008 to July 31, 2008. Table III shows the final results of the experiment. The numerical results for ISO-NE and WNN are obtained from \cite{9-guan2012very}, while the numerical results for WT-ELM-MABC are obtained from \cite{10-li2015short}. From Table III, we can see that both the single model and ensemble model are better than ISO-NE, WNN and WT-ELM-MABC. Specifically, compared with WT-ELM-MABC, the single model improves the MAPE by 16\% and MAE by 13\%. The integration model improves the MAPE by 20\% and MAE by 16\%.

\begin{table}[]
\caption{One-hour ahead forecasting MAPE(\%) and MAE of proposed method and other methods on the ISO-NE dataset. + represents the results of the ensemble method.
}
\centering
\begin{tabular}{@{}lrr@{}}
\toprule
 & \multicolumn{1}{c}{MAPE} & \multicolumn{1}{c}{MAE} \\ \midrule
ISO-NE & 0.81 & 138 \\
WNN & 0.49 & 84 \\
WT-ELM-MABC & 0.45 & 74.41 \\
Proposed & 0.38 & 64.75 \\
$Proposed^{+}$ & \textbf{0.36} & \textbf{62.43} \\ \bottomrule
\end{tabular}
\label{table3}
\end{table}

We also compare the proposed method with the other three methods \cite{36-pramono2019deep,37-tian2018deep,35-kong2017short}. The range of the training set is from 2004 to 2005, with  the  last  month  being  used  as  the  validation  set. The range of the test set is May 2006. Table IV shows the results of the experiment, the numerical results of other methods come from \cite{36-pramono2019deep}. From Table IV, we can see that our method is still better than other methods. Specifically, compared to Pramono et al \cite{36-pramono2019deep}, the single model improves MAPE by 23.91\%, MAE by 22.29\%, and RMSE by 20.81\%. The integrated model improves MAPE by 28.26\%, MAE by 27.88\%, and RMSE by 31.05\%.

\begin{table}[]
\caption{One-hour ahead forecasting MAPE(\%) and MAE of proposed method and other methods on the ISO-NE dataset. + represents the results of the ensemble method.
}
\centering
\begin{tabular}{@{}lccr@{}}
\toprule
                 & MAPE       & MAE            & RMSE           \\ \midrule
Tian et al   & 0.66          & 89.07          & 141.97         \\
Kong et al   & 0.48          & 65.12          & 100.50         \\
Wavenet       & 0.57          & 78.02          & 125.11         \\
Pramono et al & 0.46          & 62.23          & 88.31          \\
Proposed         & 0.35          & 48.36          & 69.05          \\
Proposed$^{+}$        & \textbf{0.33} & \textbf{44.88} & \textbf{60.89} \\ \bottomrule
\end{tabular}
\end{table}

\subsection{Performance of the Proposed Model on the NAU Dataset}

In this case, we compare performance of the proposed model on the NAU dataset with the existing 5 methods . In \cite{27-deihimi2012application}, ESN is applied to power load forecasting. In \cite{28-reis2005feature}, the discrete wavelet transform is embedded into the neural network for short-term load prediction. In \cite{29-amjady2009short}, the load data are decomposed through wavelet transform, and each component is predicted by combining a neural network and an evolutionary algorithm. In \cite{30-ceperic2013strategy}, particle swarm optimization is used for SVR superparameter optimization, and a parallel model consisting of 24 sets of support vectors is used for day-ahead load prediction.

The training set ranges from January 1, 1988 to October 12, 1990, with the last month being used as the validation set. The test set covers the period from October 12, 1990 to October 12, 1992. We also perturb the temperature in the original data, and we only perturb the data in the training set. As suggested in \cite{27-deihimi2012application}, Gaussian noise with a mean value of zero and standard deviation of 0.6 is added to the actual temperature data. The experimental results are shown in Table V. In the actual temperature dataset, the effect of the single model is consistent with that of WT-ELM-MABC. The single model is slightly better than WT-ELM-MABC in predicting temperature with noise. Our ensemble model achieves the best results for both the real and noisy temperatures. We also notice that the effect of the ensemble model does not change for either the actual temperatures or the noisy temperatures. In fact, the final test MAEs of the ensemble model for actual temperatures and noisy temperatures are 14.431 and 14.436, respectively, and there is little difference between the two results. This shows that the proposed model is robust to temperature noise.

\begin{table}[]
\caption{One-hour ahead forecasting MAPE(\%) of proposed method and other methods on the NAU dataset. + represents the results of the ensemble method.}
\centering
\begin{tabular}{@{}lcc@{}}
\toprule
 & \begin{tabular}[c]{@{}c@{}}Actual \\ Temperature\end{tabular} & \begin{tabular}[c]{@{}c@{}}Noisy\\ Temperature\end{tabular} \\ \midrule
ESN & 1.14 & 1.21 \\
M2 & 1.10 & 1.11 \\
WT-NN-EA & 0.99 & - \\
SSA-SVR & 0.72 & 0.73 \\
WT-ELM-MABC & 0.67 & 0.69 \\
Proposed & 0.67 & 0.68 \\
$Proposed^{+}$ & \textbf{0.64} & \textbf{0.64} \\ \bottomrule
\end{tabular}
\end{table}

\subsection{Performance between Proposed Model and Machine Learning Models}
In this case, we compare the proposed model with common machine learning models. Specifically, we compare the proposed model with random forest, gradient boosting decision tree (GBDT) \cite{31-friedman2001greedy}, Xgboost \cite{32-chen2015xgboost} and Catboost \cite{33-dorogush2018catboost} on the NAU dataset. Our training set covers 1988-1989, with the last month as the validation set. The scope of the test set is 1990. In the interest of fairness, all the models use the same input. For the machine learning models, after adjusting the superparameters with the verification set, we also use the verification set for the final model training. As shown in Table VI, our single model improves the MAPE by 72\% and MAE by 45\% relative to those of Catboost. Our integration model improves the MAPE by 76\% and MAE by 47\%. Compared with current machine learning algorithms, our model has better generalization ability.

\begin{table}[]
\caption{One-hour ahead forecasting MAPE(\%) and MAE of proposed method and machine learning methods on the NAU dataset. + represents the results of the ensemble method.}
\centering
\begin{tabular}{@{}lrr@{}}
\toprule
 & \multicolumn{1}{c}{MAPE} & \multicolumn{1}{c}{MAE} \\ \midrule
Random forest & 3.23 & 71.10 \\
GBDT & 1.98 & 45.35 \\
Xgboost & 1.34 & 30.16 \\
Catboost & 1.24 & 29.43 \\
Proposed & 0.72 & 16.28 \\
$Proposed^{+}$ & \textbf{0.69} & \textbf{15.59} \\ \bottomrule
\end{tabular}
\end{table}

\subsection{Robustness Analysis of the Proposed Model}
In the actual deployment environment, there will be a slight deviation between the final acquired value and the actual actual value due to an error in the measurement equipment or an error in the value recording. This fact indicates that the model applied to short-term load forecasting should have good robustness, that is, the slight disturbance of the input by the model should not cause excessive deviation of the output. To ensure the reliability of the proposed model prediction, we perturb the load data and the temperature data to varying degrees to explore the robustness of the model. Specifically, we use a Gaussian distribution with a mean of 0 and a standard deviation of (0, 0.3, 0.6, 0.9, 1.2, 1.5, 1.8, and 2.1) to generate eight sets of noise. The maximum range of the generated noise is [-8.0, 10.57]. Then, these 8 sets of noise are added to the load data and temperature data. We only perturb the training set. We conduct the experiment using the ISO-NE dataset, with the training set ranging from January 2007 to June 2008 and the last month as the validation set. The scope of the test set is July 2008. The experimental results are shown in Figure 5. We find that there is no significant difference between the MAPEs of all the models, even when the temperature and load data are disturbed to different degrees. Interestingly, the MAPE of the model does not increase when the noise of the load value and temperature value increases. Conversely, adding noise improves the MAPE of the model. For example, when the perturbation variance of the temperature is 0 and the perturbation variance of the load value is 0.3, the MAPE is lower than that of the undisturbed model. The experimental results show that the proposed model is robust to data noise.

\begin{figure}
\centering
\includegraphics[ width=0.38\textwidth]{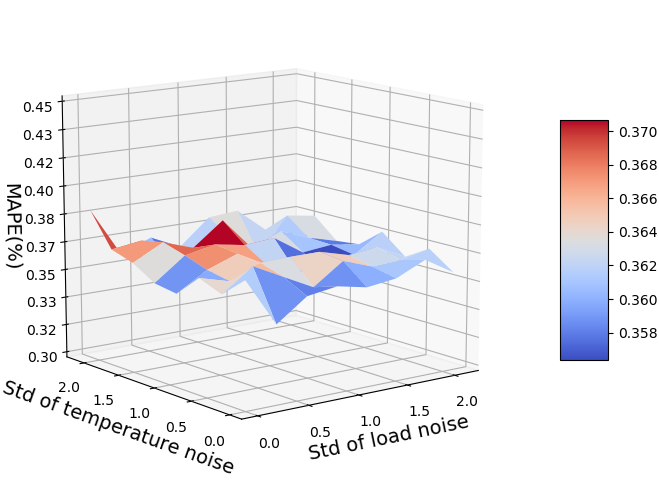}
\caption{MAPEs(\%) of models with varying degrees of perturbation of temperature and load data. The perturbation data is generated by Gaussian distribution with a mean value of 0 and a variance of (0, 0.3, 0.6, 0.9, 1.2, 1.5, 1.8, and 2.1).}
\end{figure}

\section{Conclusion}
In this paper, we first propose the dense average connection. Dense average connection can solve the problem of gradient explosion well  which makes it possible to build a deep model. Based on the dense average connection, we build the dense average network for load forecasting. Compared with existing methods on two public datasets, dense average network has better prediction effect. We also find that the ensemble model can reduce the standard deviation and peak value of the final prediction bias compared to a single model. To verify the reliability of the model predictions, we also disturb the input of the model to different degrees. The experimental results show that the proposed model has good robustness.

In Section III, we found that the appropriate disturbance of load and temperature data will not significantly reduce the prediction performance, or even prompt the prediction effect. In the future work, we will explore whether data disturbance can be used as a way of data enhancement to further improve the effect of load forecasting.

\ifCLASSOPTIONcaptionsoff
  \newpage
\fi







\end{document}